\documentclass[10pt,twocolumn,letterpaper]{article}

\usepackage{cvpr}
\usepackage{times}
\usepackage{epsfig}
\usepackage{graphicx}
\usepackage{amsmath}
\usepackage{amssymb}
\usepackage{helvet}  
\usepackage{courier} 
\usepackage{url} 
\usepackage{multirow}
\usepackage{bm}
\usepackage{amsfonts}
\usepackage{subcaption}

\usepackage[pagebackref=true,breaklinks=true,letterpaper=true,colorlinks,bookmarks=false]{hyperref}

\cvprfinalcopy 

\def\cvprPaperID{164} 

\ifcvprfinal\fi
\begin{document}

\title{Robustness Analysis of Visual QA Models by Basic Questions}

\author{Jia-Hong Huang~~~~~~~Cuong Duc Dao*~~~~~~~Modar Alfadly*~~~~~~~C. Huck Yang~~~~~~~Bernard Ghanem\\
King Abdullah University of Science and Technology ; Georgia Institute of Technology\\
{\tt\small \{jiahong.huang, dao.cuong, modar.alfadly, bernard.ghanem\}@kaust.edu.sa}
\tt\small ; huckiyang@gatech.edu
}

\maketitle

\begin{abstract}
Deep neural networks have been playing an essential role in many computer vision tasks including Visual Question Answering (VQA). Until recently, the study of their accuracy has been the main focus of research and now there is a huge trend toward assessing the robustness of these models against adversarial attacks by evaluating the accuracy of these models under increasing levels of noisiness. In VQA, the attack can target the image and/or the proposed main question and yet there is a lack of proper analysis of this aspect of VQA. In this work, we propose a new framework that uses semantically relevant questions, dubbed basic questions, acting as noise to evaluate the robustness of VQA models. We hypothesize that as the similarity of a basic question to the main question decreases, the level of noise increases. So, to generate a reasonable noise level for a given main question, we rank a pool of basic questions based on their similarity with this main question. We cast this ranking problem as a $LASSO$ optimization problem. We also propose a novel robustness measure $R_{score}$ and two large-scale question datasets, General Basic Question Dataset and Yes/No Basic Question Dataset in order to standardize robustness analysis of VQA models. We analyze the robustness of several state-of-the-art VQA models and show that attention-based VQA models are more robust than other methods in general. The main goal of this framework is to serve as a benchmark to help the community in building more accurate \emph{and} robust VQA models.
\end{abstract}

\section{Introduction}

\vspace{3pt}\noindent\textbf{Motivations.~}

Visual Question Answering (VQA) is one of the most challenging computer vision tasks in which an algorithm is given a natural language question about an image and tasked with producing a natural language answer for that question-image pair. Recently, various VQA models \cite{4,5,9,31,37,41,57,58,59,77,82} have been proposed to tackle this problem, and their main performance measure is accuracy. However, the community has started to realize that accuracy is not the only metric to evaluate model performance. More specifically, these models should also be robust, \ie, their output should not be  affected much by some small \emph{noise} or \emph{perturbation} to the input. 
The idea of analyzing model robustness as well as  training robust models is already a rapidly growing research topic for deep learning models applied to images \cite{61,62,63}. However, and to the best of our knowledge,  an acceptable and standardized method to measure  robustness in VQA models does not seem to exist. As such, this paper is the first work to analyze  VQA models from this point of view by proposing a robustness measure and a standardized large-scale dataset.

\begin{figure}[t]
\begin{center}
   \includegraphics[width=0.98\linewidth]{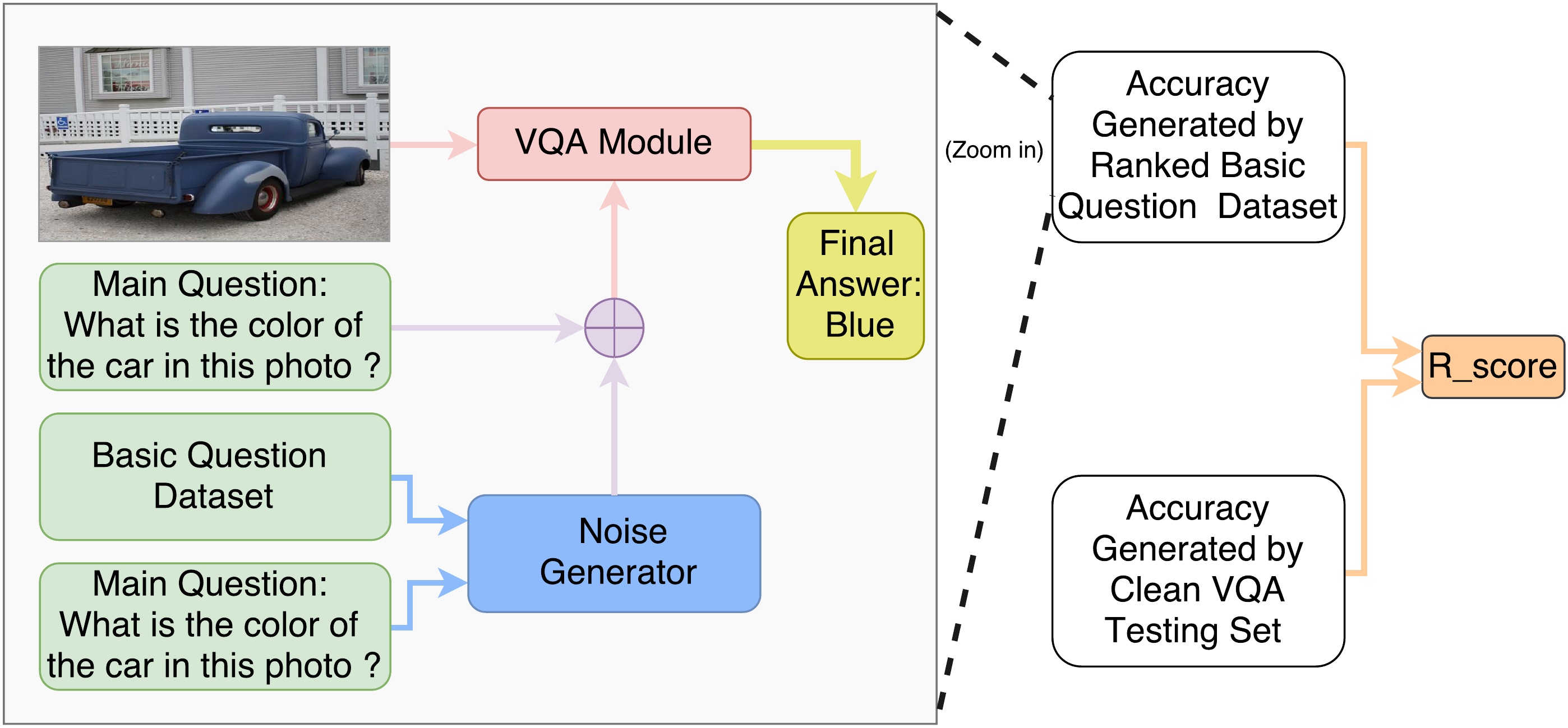}
\end{center}
   \caption{Our proposed framework for measuring the robustness of VQA models. The $R_{score}$ -- our proposed robustness measure -- is generated by the two white boxes. In the upper white box, we have two main components, VQA Module and Noise Generator, and the detail of the noise generator can be referred to Figure \ref{fig:figure2}. ``$\oplus$'' denotes the direct concatenation of basic questions.}
\vspace{-0.5cm}
\label{fig:figure1}
\end{figure}


\begin{figure}[t]
\begin{center}
   \includegraphics[width=0.98\linewidth]{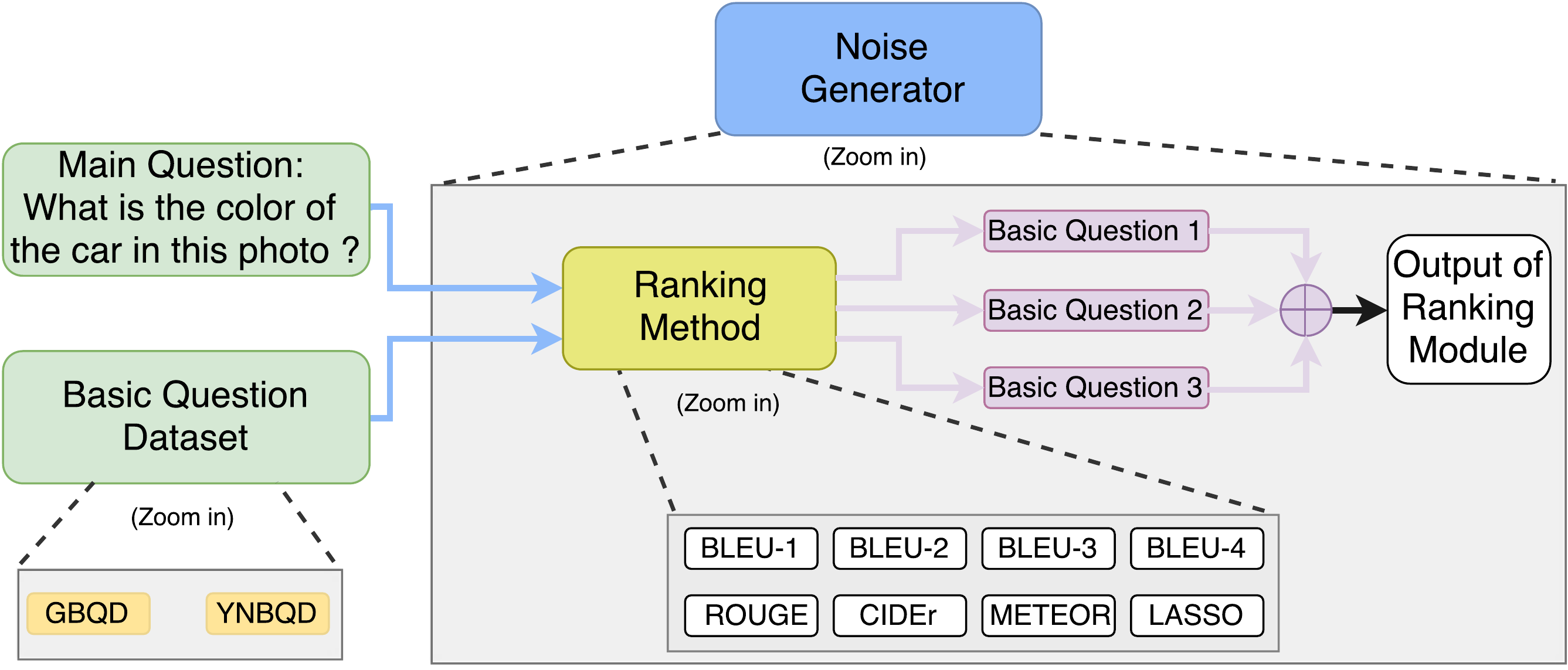}
\end{center}
   \caption{The figure shows details of Noise Generator. We have two choices, GBQD and YNBQD, of Basic Question Dataset and eight choices, BLEU-1, BLEU-2, BLEU-3, BLEU-4, ROUGE, CIDEr, METEOR and LASSO, of questions ranking methods. If a new Basic Question Dataset or ranking method is proposed in the future, we will also add them into our proposed framework. The output of Noise Generator is the concatenation of three ranked basic questions. ``$\oplus$'' denotes the direct concatenation of basic questions.}
\vspace{-0.5cm}
\label{fig:figure2}
\end{figure}

\vspace{3pt}\noindent\textbf{Assumptions.~} 

\begin{figure}[t]
\begin{center}
\includegraphics[width=0.98\linewidth]{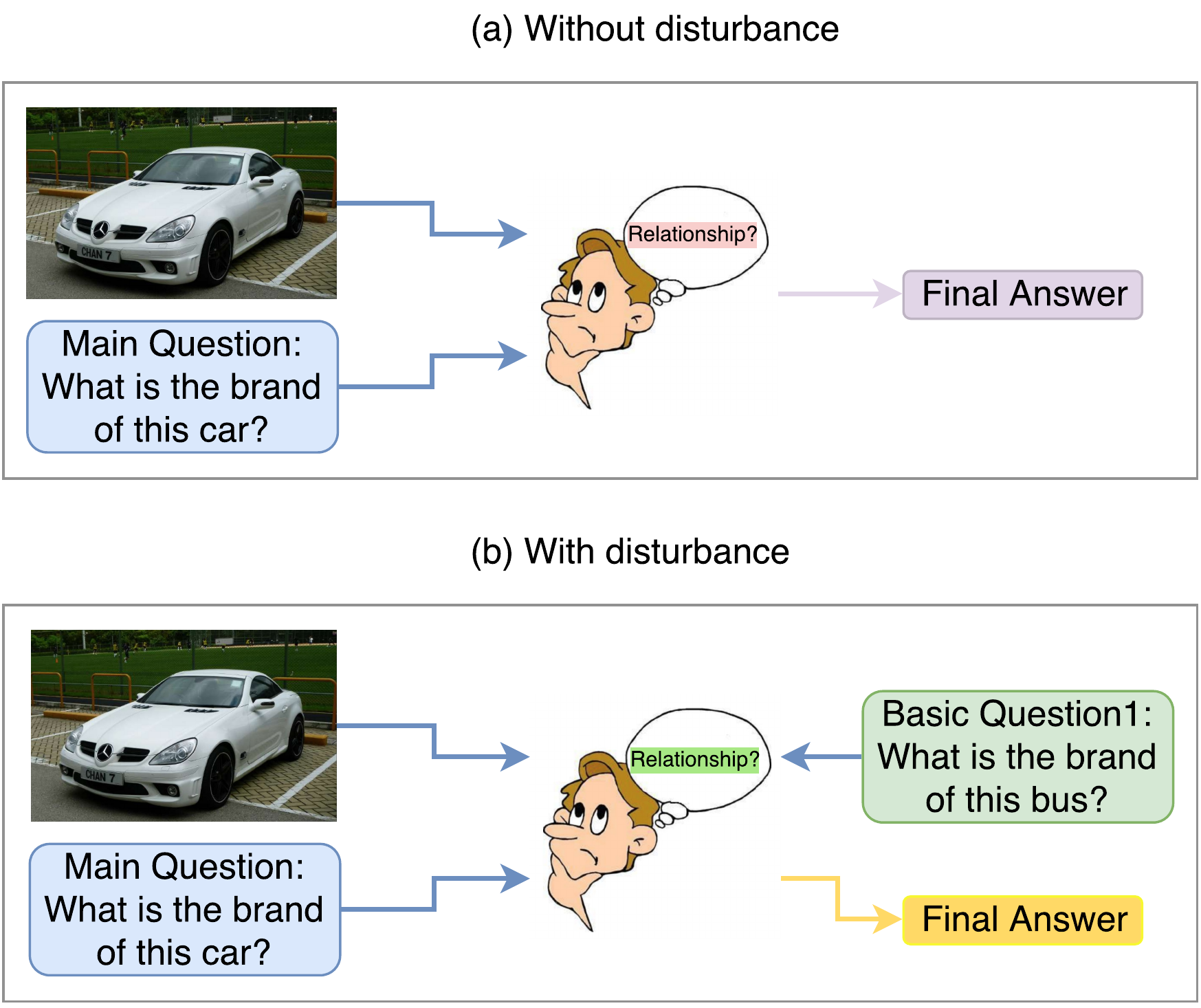}
\end{center}
\vspace{-0.30cm}
   \caption{Inspired by Deductive Reasoning in Human Thinking \cite{74}, this figure showcases the behavior of humans when subjected to multiple questions about a certain subject. Note that the relationships and the final answer in the case (a) and (b) can be different.
}
\vspace{-0.5cm}
\label{fig:figure3}
\end{figure}

The ultimate goal is for VQA models to perform as humans do for the same task. If a human is presented with a question or this question accompanied with some highly similar questions to it, he/she tends to give the same or a very similar answer in both cases. Evidence of this has been reported on in the psychology domain. Therefore, when we add or replace some words or phrases by similar words or phrases to the  query question, called the main question, the VQA model should output the same or a very similar answer. In some sense, we consider similar words or phrases as small perturbations or noise to the input, so we say that the model is robust if it produces the same answer. Note that we define a basic question as a semantically similar question to the given main question. Based on evidence from deductive reasoning in human thinking \cite{74}, we consider  basic questions as noise. In Figure \ref{fig:figure3}, cases (a) and (b) explain the general idea. In case (a), the person may have the answer ``Mercedes Benz'' in mind. However, in case (b), he/she would start to think about the relations among the two given questions and candidate answers to form the final answer which may be different from the final answer in case (a). If the person is given more basic questions, he/she would start to think about all the possible relations of all the provided questions and possible answer candidates. These relationships will clearly be more complicated, especially when the additional basic questions have low similarity score to the main question. In such cases, they will mislead the person. That is to say, those extra basic questions are large disturbances in some sense. Because robustness analysis requires studying the accuracy of VQA models under different noise levels, we need to know how to quantify the level of noise for the given question. We hypothesize that a basic question with larger similarity score to the main question is considered to inject a smaller amount of noise if it is added to the main question and vice versa. Our proposed $LASSO$ basic question ranking method is one way to quantify and control the strength of this injected noise level.

\section{Robustness Framework}

Inspired by the above reasoning, we propose a novel framework for measuring the robustness of VQA models. Figure \ref{fig:figure1} depicts the structure of our framework. It contains two modules, a VQA model and a Noise Generator. The Noise Generator, illustrated in Figure \ref{fig:figure2}, takes a plain text main question (MQ) and a plain text basic question dataset (BQD) as input. It starts by ranking the basic questions in BQD by their similarity to MQ using some text similarity ranking method. Then, depending on the required level of noise, it takes the top $n$ (\eg, $n=3$) ranked BQs and directly concatenates them. The concatenation of these BQs  with MQ is the generated noisy question. Instead of feeding the MQ to the VQA model, we  replace it with the generated noisy question and measure the accuracy of the output. To measure the robustness of this VQA model, the accuracy with and without the generated noise is compared. To this end, we propose a robustness measure  $R_{score}$ to quantify this comparison. 

For the questions ranking method \cite{65,66}, given any two questions we can have different measures that quantify the similarity of those questions and produce a score between $[0-1]$. Using such similarity measures, we can have different rankings of the similarity of MQ to the questions in BQD, where the BQs with higher similarity score to MQ  rank higher than those with less similarity. Along those lines, we propose a new question ranking method formulated using $LASSO$ optimization and compare it against other rankings produced by seven different yet popular textual similarity measures. We do this comparison to rank our proposed BQDs, General Basic Question Dataset (GBQD) and Yes/No Basic Question Dataset (YNBQD). Furthermore, we evaluate the robustness of six pretrained state-of-the-art VQA models \cite{4,41,57,59}. Finally, extensive experiments show that $LASSO$ is the best BQD ranking method among others.

\vspace{3pt}\noindent\textbf{Contributions.~} 

\textbf{(i)} We propose a novel framework to measure the robustness of VQA models and test it on six different models.
\textbf{(ii)} We propose a new text-based similarity ranking method and compare it against seven popular similarity metrics, BLEU-1, BLEU-2, BLEU-3, BLEU-4 \cite{49}, ROUGE \cite{68}, CIDEr \cite{67} and METEOR \cite{69}. Then, we show that our $LASSO$ ranking method is the best among them.
\textbf{(iii)} We introduce two large-scale basic questions datasets: General Basic Question Dataset (GBQD) and Yes/No Basic Question Dataset (YNBQD).
\noindent

\section*{Acknowledgement}

This work is supported by competitive research funding from King Abdullah University of Science and Technology (KAUST) and the High Performance Computing Center in KAUST.



{\small
\bibliographystyle{unsrt}
\bibliographystyle{ieee}
\bibliography{egbib}
}

\end{document}